\documentclass[letterpaper]{article} 
\usepackage{aaai24}  
\usepackage{times}  
\usepackage{helvet}  
\usepackage{courier}  
\usepackage[hyphens]{url}  
\usepackage{graphicx} 
\usepackage{natbib}  
\usepackage{caption} 
\usepackage{algorithm}
\usepackage{algorithmic}

\usepackage{newfloat}
\usepackage{listings}
\DeclareCaptionStyle{ruled}{labelfont=normalfont,labelsep=colon,strut=off} 
\floatstyle{ruled}
\newfloat{listing}{tb}{lst}{}
\floatname{listing}{Listing}
\usepackage{color} 
\usepackage{multirow}
\nocopyright 

\title{Empirical Evidence for the Fragment level Understanding \\ on Drug Molecular Structure of LLMs}
\author{
    Xiuyuan Hu\textsuperscript{\rm 1}, Guoqing Liu\textsuperscript{\rm 2}, Yang Zhao\textsuperscript{\rm 1}, Hao Zhang\textsuperscript{\rm 1}\thanks{Corresponding author: Hao Zhang. \\ Accepted by AAAI 2024 workshop: Large Language Models for Biological Discoveries (LLMs4Bio).}
}
\affiliations{
    \textsuperscript{\rm 1}Department of Electronic Engineering, Tsinghua University, Beijing, China, \textsuperscript{\rm 2}Microsoft Research AI4Science\\
    huxy22@mails.tsinghua.edu.cn, guoqingliu@microsoft.com, \{zhao-yang,haozhang\}@tsinghua.edu.cn
}

\usepackage{bibentry}

\begin{document}

\maketitle

\begin{abstract}
AI for drug discovery has been a research hotspot in recent years, and SMILES-based language models has been increasingly applied in drug molecular design. However, no work has explored whether and how language models understand the chemical spatial structure from 1D sequences. In this work, we pre-train a transformer model on chemical language and fine-tune it toward drug design objectives, and investigate the correspondence between high-frequency SMILES substrings and molecular fragments. The results indicate that language models can understand chemical structures from the perspective of molecular fragments, and the structural knowledge learned through fine-tuning is reflected in the high-frequency SMILES substrings generated by the model.
\end{abstract}

\section{1~~~Introduction}

With the explosion of computational technologies such as deep learning and the accumulation of data across various disciplines including biology, chemistry, medicine, and pharmacology, the potential of artificial intelligence (AI) in drug discovery has increasingly attracted attention in recent years \cite{MLDDSurvey, AIDDSurvey}. Compared to traditional wet-laboratory-based drug research and development processes, computer-aided drug discovery (CADD) has evident advantages in speed and cost, exemplified by techniques such as high-throughput screening \cite{CADDReview1}. Notably, AI tools 
may drive a greater pharmaceutical revolution, as they not only further improve computational accuracy and efficiency but also demonstrate unprecedented creativity \cite{AI4ScienceSurvey}.

Large language models (LLMs) based on transformer architecture have swept the field of natural language processing (NLP), especially achieving stunning breakthroughs in understanding and generation abilities for conversational intelligence \cite{InstructGPT, GPT4Report, LLMsSurvey}. However, the performance of the currently best general language model, GPT-4, is still limited in the domain of drug discovery; for example, its reliability in accomplishing quantitative tasks and processing molecular sequences is doubtful \cite{GPT4SciSurvey}. As it stands, training specialized language models for chemical language to deal with problems in drug discovery seems more meaningful than using general language models.

At present, generative language models have been widely applied in drug molecular design, and these models can generate a set of desirable molecules according to given objectives. Particularly, existing works have employed transformer-based language models to process chemical molecular strings; examples include MolGPT \cite{MolGPT}, Chemformer \cite{Chemformer} and MolRL-MGPT \cite{MolRL-MGPT}. For this reason, tokenizers specifically designed for chemical languages have been introduced, and models are trained on various molecular data. The chemical language most commonly used by current AI researchers is the simplified molecular input line entry system (SMILES) \cite{SMILES}, which can represent any molecular structure by a sequence of characters.

However, although SMILES-based drug design approaches are competitive \cite{MolOpt}, understanding and generating 1D molecular strings appear less intuitive and less interpretable than methods based on 2D molecular graphs. For instance, two atoms connected by a chemical bond in a molecule may not be adjacent in a SMILES string, posing a challenge for language models to correctly comprehend chemical structures. Therefore, to design better and more reliable LLMs for drug discovery, it is necessary to reveal whether and how language models understand the chemical structure of molecules.

To address this problem, we follow the pre-training and fine-tuning paradigm of LLMs: pre-training a transformer language model on chemical molecular data and then fine-tuning this model using reinforcement learning for drug molecular design. We carry out experiments on three design tasks, not only validating the effectiveness of the algorithm but also studying changes in the model's understanding of chemical language during the fine-tuning process. Specifically, we analyze the learning of molecular spatial structures by the language model from the perspective of molecular fragments. We demonstrate an increase in the quantity and quality of high-frequency SMILES substrings generated by the language model during the fine-tuning process and point out that these substrings correspond well with substructures in the 2D molecular diagrams. These results indicate that the language model for molecular generation understands the spatial structure of molecules, rather than merely fitting the SMILES sequences. Our code is available at: \url{https://github.com/HXYfighter/LLMsUnderstandMol}.

\section{2~~~Related Works}

In recent years, pharmacology has attracted increasing attention, especially as global public health crises such as the COVID-19 pandemic have demanded greater efficiency in drug development. Consequently, researchers worldwide are committed to applying advanced AI technologies to drug discovery, aiming to tackle practical problems such as drug-target binding prediction, molecular property prediction, retrosynthesis, and drug molecular design \cite{MLDDSurvey2}. Although algorithms have seen some performance improvements, they still fall short of meeting the demands of real-world drug discovery, one key challenge being their lack of interpretability.

Drug molecular design is a generative task with the objective of generating a diverse set of drug candidates that fulfill given properties. Given that the chemical space is unstructured and extremely vast \cite{EstimateChemSpace}, drug molecular design is quite challenging. Machine learning techniques have been extensively utilized to address this issue, including genetic algorithms \cite{GraphGA,ReinGA}, reinforcement learning \cite{GCPN,MolDQN,PGFS,GEGL,MOLER}, generative adversarial networks \cite{ORGAN}, flow models \cite{MoFlow,GFlowNet}, etc. Common representations for small chemical molecules encompass 1D molecular strings, 2D molecular graphs, and 3D geometries, among which 1D molecular strings are the most suitable for processing and generation using language models \cite{MolGenSurvey}.

\subsection{2.1~~Language Models for Drug Design}

Significant advancements in the NLP field in recent years have driven rapid updates of drug design methods based on 1D string representation. Currently, the most commonly used 1D string molecular representation is SMILES \cite{SMILES}, characterized by its non-necessarily valid strings and the possibility of a single molecular structure to correspond to multiple different SMILES strings. Besides, other 1D sequence molecular representations, including SELFIES \cite{SELFIES} and IUPAC \cite{IUPAC}, are also utilized in drug discovery algorithms.

Reinvent \cite{Reinvent} was the first to propose the use of reinforcement learning to fine-tune a recurrent neural network (RNN) to accomplish SMILES-based drug design, and it continues to lead in molecular design benchmarks to date \cite{MolOpt}. Subsequently, numerous SMILES-based molecular generation methods were proposed, such as ReLeaSE \cite{ReLeaSE} and ChemGE \cite{SMILES-GA}, among others. In addition, some algorithms utilize transformer models for molecule generation, including MCMG \cite{MCMG}, MolGPT \cite{MolGPT}, Chemformer \cite{Chemformer}, GMTransformer \cite{GMTransformer}, FSM-DDTR \cite{FSM-DDTR}, and MolRL-MGPT \cite{MolRL-MGPT}.

While much research is dedicated to uncovering the working mechanisms of LLMs, no study has focused on explaining chemical language models. We believe it is necessary to meticulously explore whether and how existing language models model and understand complex relationships between molecular structures and properties. This exploration holds significant implications for further tapping into the potential of language models in drug discovery.

\subsection{2.2~~Fragment-based Drug Design}

Fragment-based drug design (FBDD) is a contemporary strategy employed in medicinal chemistry for the discovery and optimization of novel drug compounds, which uses smaller chemical compounds or 'fragments' as starting points in the creation of new drugs. This method offers several potential advantages over traditional ones, including (1) better coverage of the chemical space with fewer molecules, (2) lower combinatorial complexity to build up compounds, and (3) higher hit rates due to the rational design process \cite{FBDDSurvey1,FBDDSurvey2}.

As a result, the concept of FBDD has also been embraced by AI researchers. For instance, HierG2G \cite{HierG2G}, RationaleRL \cite{RationaleRL}, FREED \cite{FREED} RS-VAE \cite{RS-VAE} and MiCaM \cite{MiCaM} propose different approaches to mine molecular substructures/motifs to generate molecules.

Up until now, no work has attempted to implement fragment-based drug design using language models, nor has any work analyzed drug design language models from the perspective of fragments. Therefore, we aim to fill this gap.
\section{3~~~Pre-training on Chemical Language}

To equip language models with the fundamental capability of processing chemical language, we pre-train them on a dataset of molecules in SMILES representation, aiming to enable the model to comprehend the chemical syntax of SMILES and generate valid SMILES strings. Following MolGPT \cite{MolGPT}, we construct an auto-regressive generative model based on GPT-2 architecture \cite{GPT-2}, which comprises only 6.4M parameters, including eight transformer blocks. For the chemical language of SMILES, we establish a vocabulary consisting of over 100 atomic-level tokens that incorporate atoms, functional groups, numbers, symbols, etc., and build a corresponding tokenizer.

\begin{figure}[ht]
    \centering
    \includegraphics[width=\linewidth]{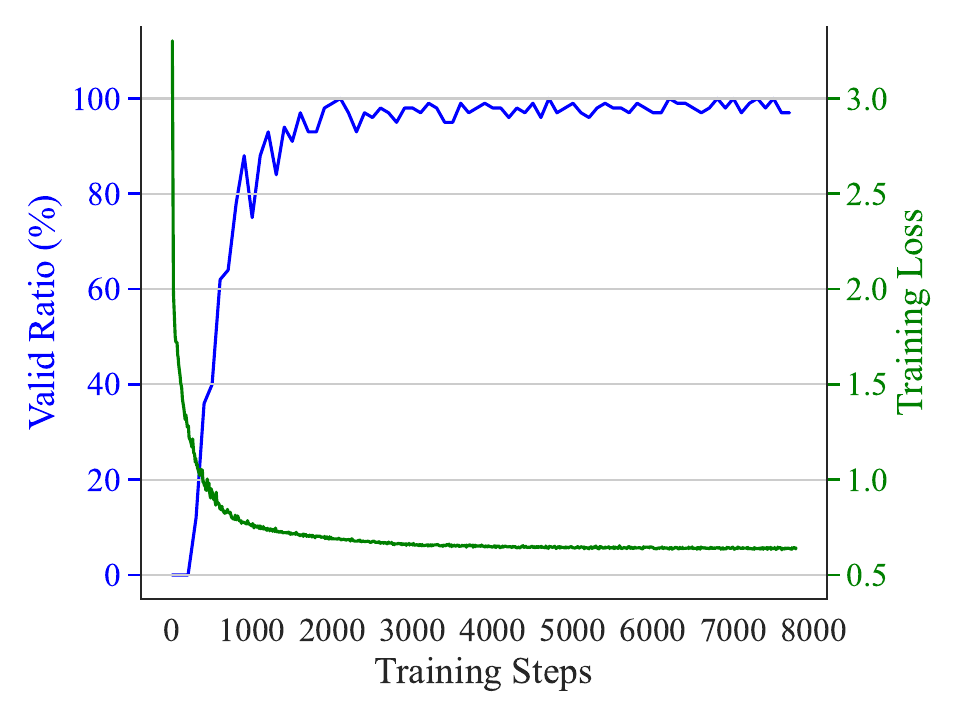}
    \caption{The changing curves of valid ratio and training loss during the pre-training process on chemical language.}
    \label{pretrain}
\end{figure}

On the ChEMBL dataset \cite{ChEMBL}, which contains over two million drug molecules, we train the aforementioned model unsupervised after filtering out some overly long SMILES strings or those containing rare ions. Specifically, we pre-train the GPT model on the dataset for 10 epochs with a batch size of 4096, and the learning rate is 0.001 with a cosine decay schedule. During the training process, we monitor the change in cross-entropy losses and the valid ratio of SMILES strings generated by the model, as shown in Figure \ref{pretrain}. The loss appears to converge, and the valid ratio is stable at higher than 97\%, demonstrating that the model is indeed gradually mastering the ability to handle chemical language.

It is worth noting that, compared with popular large models in the NLP field, our model is not "large". This is because the number of tokens in the chemical language is far fewer than in natural language, and the length of one "sentence", i.e., a SMILES string, is relatively limited (usually dozens of tokens). Therefore, a smaller model is enough to master chemical syntax without needing a large model with billions of parameters.
\section{4~~~Fine-tuning for Drug Molecular Design}

\subsection{4.1~~RL Fine-tuning}
In the task of drug molecular design, once we have obtained a language model capable of generating valid SMILES strings, we hope to fine-tune it to generate molecules that exhibit specific properties, i.e., drug candidates. We use scoring functions (also known as oracles) to evaluate the properties of molecules, such as using docking software to predict whether a molecule can bind well with a given protein target. Generally speaking, there are virtually no compounds satisfying conditions (i.e., \textit{de novo} drug design) in existing datasets, so supervised learning for fine-tuning is not feasible, while reinforcement learning (RL) is currently the most recognized solution for this problem. 
In an RL fine-tuning process for drug molecular design, the language model is the agent to be fine-tuned; its auto-regressive generation of SMILES tokens represents RL actions, and the scores of molecular properties are RL rewards. 

The transformer model pre-trained on the SMILES dataset is adopted to initialize our RL agent. Following Reinvent \cite{Reinvent}, we use REINFORCE algorithm \cite{REINFORCE} to fine-tune the agent at each step with the following loss function:
\begin{equation}
L(x)=[\log P(x, \Theta_{\mathrm{pre}})-\log P(x, \Theta_{\mathrm{agent}})+\sigma\cdot s(x)]^2
\end{equation}
where $x="x^0x^1...x^n"$ is a sequence of tokens generated by the agent, $\Theta_{\mathrm{pre}}$ and $\Theta_{\mathrm{agent}}$ respectively represents the weights of the pre-trained chemical language model and the weights of the agent, $s(\cdot)$ refers to the scoring function given by the drug design task, and $\sigma$ is a coefficient hyper-parameter for controlling the term of scores. The likelihood $P$ of generating the sequence $x$ equals the product of the probabilities of generating each token in turn:
\begin{equation}
P(x, \Theta)=\prod_{i=1}^np(x^i,\Theta|x^0,...,x^{i-1})
\end{equation}

\begin{algorithm}[tb]
\caption{RL fine-tuning algorithm}
\label{RLalgorithm}
\textbf{Input}: \\
(1) Pre-trained weights $\Theta_{\mathrm{pre}}$, (2) Scoring function $s(\cdot)$ \\
\textbf{Hyper-parameters}: \\
(1) Number of fine-tuning steps $t$, (2) Batch size $m$, \\
(3) Learning rate $\eta$, (4) Coefficient $\sigma$ \\
\textbf{Output}: \\
A set of SMILES strings of high-scoring molecules \\
\textbf{Algorithm flow}:
\begin{algorithmic}[1] 
\STATE Create an empty memory of high-scoring molecules $M$
\STATE Initialize an agent $\Theta_{\mathrm{agent}}$ with $\Theta_{\mathrm{pre}}$
\FOR{$i=1,2,...,t$}
\STATE The agent samples a batch of strings $x_1,...,x_m$
\STATE Using $s(\cdot)$ to score the strings
\STATE Update $M$ with pairs $(x_k,s(x_k)),\ k=1,...m$
\STATE Calculate losses $L(x_k),\ k=1,...m$
\STATE Update $\Theta_{\mathrm{agent}}$ using losses with the learning rate $\eta$
\ENDFOR 
\STATE \textbf{return} $M$
\end{algorithmic}
\end{algorithm}

\begin{figure*}[ht]
    \centering
    \includegraphics[width=\linewidth]{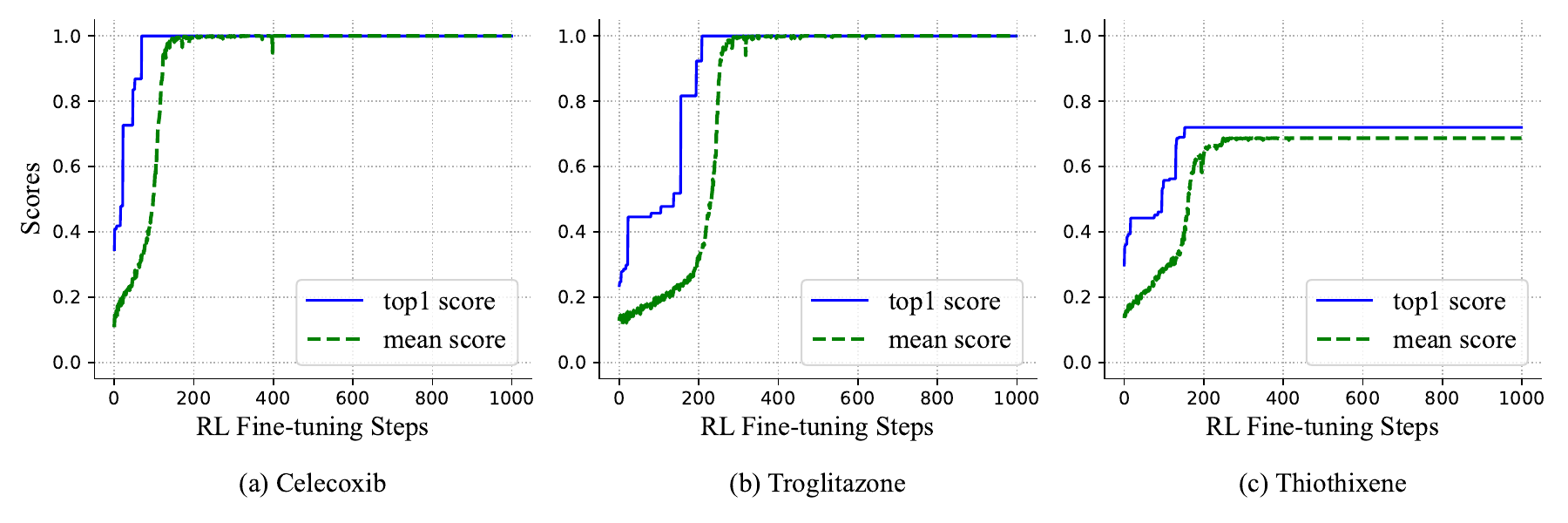}
    \caption{The score curves during the RL fine-tuning processes on three drug design tasks.}
    \label{finetune_scores}
\end{figure*}

\subsection{4.2~~Experiments}

\begin{figure}[ht]
    \centering
    \includegraphics[width=0.8\linewidth]{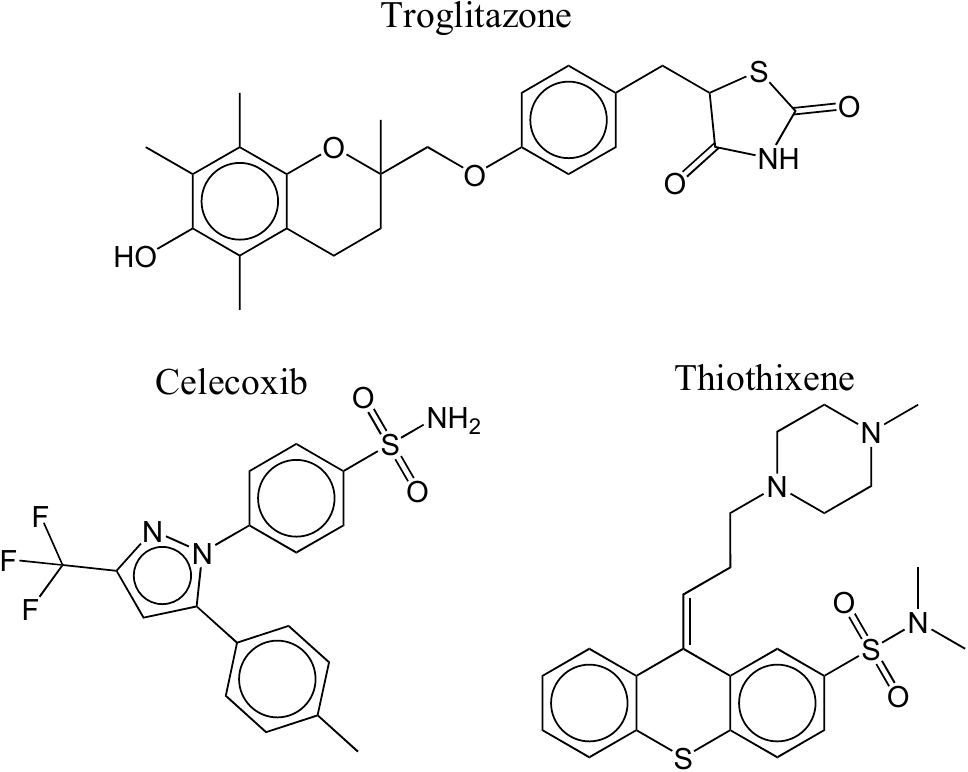}
    \caption{Three target drug structures for rediscovery tasks.}
    \label{Drugs}
\end{figure}

For experiments, we select three drug rediscovery tasks from the GuacaMol benchmark \cite{GuacaMol} as fine-tuning objectives. The target drug compounds are Celecoxib, Troglitazone, and Thiothixene, respectively (as shown in Figure \ref{Drugs}). The scoring function calculates the Tanimoto similarity \cite{Tanimoto} between the molecule corresponding to a generated SMILES string and the target structure, with values ranging from 0 to 1, where a score of 1 signifies successful discovery of the target structure; invalid strings produced receive a score of -1.

For each design objective, we perform 1000 steps of RL fine-tuning, where the batch size sampled in each step is 256, the learning rate is set to 0.0001, and $\sigma$ is set to 1000.

Figure \ref{finetune_scores} demonstrates the score curves of the SMILES strings generated by the agent during the RL fine-tuning process. This includes the average score of all strings sampled at each step (\textcolor[rgb]{0, 0.6, 0}{mean score}) and the highest score among the previously discovered molecules (\textcolor{blue}{top-1 score}), which is the official drug design evaluation metric in GuacaMol benchmark. Results show that in the three drug rediscovery tasks, the top-1 scores achieve a full score of 1.0 for both Celecoxib and Troglitazone, indicating successful completion of the task objectives. Although Thiothixene is not successfully rediscovered, the top-1 score reaches 0.72, demonstrating that RL fine-tuning has an effective impact.

Moreover, the mean scores of the sequences produced by the GPT agent also increase during the RL fine-tuning process and steadily converged to a level close to the top-1 scores. This suggests that the chemical knowledge acquired by the language model indeed transfer from a relatively broad distribution to a small range close to the drug design objectives. Furthermore, this RL-based transferring strategy proved to be applicable for various objectives.
\section{5~~~Language Models' Understanding of Molecular Fragments}

In this section, we aim to explore the language models' understanding of chemical structures. Specifically, we seek to verify whether language models, through task-oriented fine-tuning, truly learn spatial structural fragments of the target molecules instead of merely fitting the SMILES sequences. To this end, we employ the SMILES pair encoding algorithm to extract high-frequency substrings from the set of SMILES strings sampled by the language model, which reflect the patterns grasped by the language model toward the task objectives. In the three tasks of drug molecular design, we study the quantity and quality changes of high-frequency substrings during the fine-tuning process; here, quality refers to their efficiency in encoding the SMILES strings of the target molecules.

\subsection{5.1~~SMILES Pair Encoding}

To analyze "molecular fragments" in chemical language, we use the SMILES Pair Encoding (SPE) algorithm \cite{SmilesPE} to identify high-frequency substrings in a set of chemical strings. Its principle is derived from the Byte Pair Encoding (BPE) algorithm \cite{BytePE}, which is commonly used in NLP and is effective in data compression and text tokenization.

Specifically, for a given set of SMILES strings, we first apply SMILES randomization (enumeration) \cite{SMILESEnumeration,SMILESRandomization} as a data augmentation, thereby ensuring that adjacency in the chemical space is also reflected in the SMILES set as far as possible. Subsequently, the SPE algorithm identifies all high-frequency substrings in this set through an iterative process. In each iteration step, the occurrence frequencies of all token pairs are counted, and the highest-frequency token pair is merged and encoded into a new token. The iterative process continues until the occurrence frequencies of all token pairs are lower than a pre-defined minimum frequency (MF) hyper-parameter.

It is worth noting that the SMILES substrings recognized by the SPE algorithm do not necessarily perfectly correspond to substructures of molecules, such as "O)", "c1", etc. This is because SMILES, as a sequence representation, incorporates numbers, parentheses, etc., to encode chemical spatial structures such as cycles. These tokens do not correspond to atoms, chemical bonds, or other chemical entities. However, like punctuation in natural language, they also play significant syntactic roles in chemical language, so they should also be included in our analysis of the chemical language model.

\begin{figure}[ht]
    \centering
    \includegraphics[width=\linewidth]{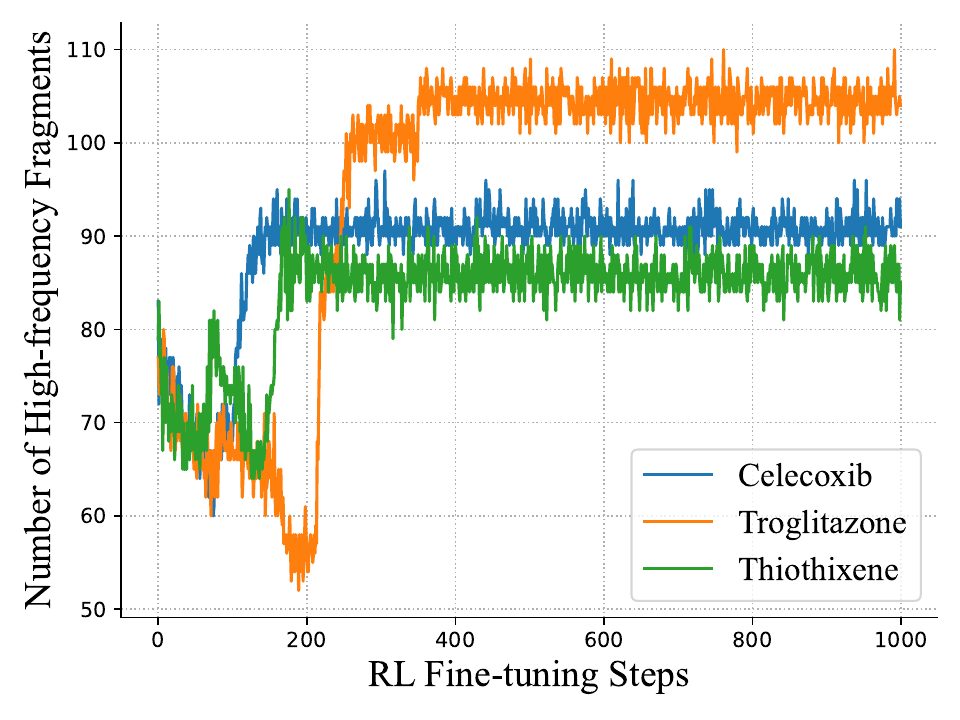}
    \caption{The changing curves of numbers of high-frequency fragments extracted by SPE algorithm during the RL fine-tuning processes on three drug design tasks.}
    \label{highfreq}
\end{figure}

\begin{table*}[htb]
\centering
\caption{The canonical and randomized SMILES strings of the three target drug structures for experimental analysis. The colored substrings correspond to the colored substructures in Figure \ref{DrugFrag}.}
\label{DrugSmiles}
\begin{tabular}{ccc}
\hline
\textbf{Target Drugs} &  & \textbf{SMILES strings} \\ \hline
\multirow{3}{*}{Celecoxib} & (canonical) & Cc1ccc(-\textcolor{red}{c2cc(C(F)(F)F)nn2-}c2ccc\textcolor{blue}{(S(N)(=O)=O)}cc2)cc1 \\
                  & rand1 & c1(-c2ccc(C)cc2)n(-c2ccc\textcolor{blue}{(S(N)(=O)=O)}cc2)nc(C(F)(F)F)c1 \\
                  & rand2 & c1c\textcolor{blue}{(S(N)(=O)=O)}ccc(\textcolor{red}{-n2nc(C(F)(F)F)cc2}-c2ccc(C)cc2)c1 \\ \hline
\multirow{3}{*}{Troglitazone} & (canonical) & Cc1c(C)c2c(c(C)c1O)CC\textcolor{red}{C(C)(COc1ccc(CC3}\textcolor{blue}{SC(=O)NC3=O)}cc1)O2 \\
                  & rand1 & \textcolor{red}{CC1(COc2ccc(CC3}\textcolor{blue}{C(=O)NC(=O)S}3)cc2)Oc2c(C)c(C)c(O)c(C)c2CC1 \\
                  & rand2 & c12c(c(C)c(O)c(C)c1C)CC\textcolor{red}{C(C)(COc1ccc(CC3}\textcolor{blue}{C(=O)NC(=O)S}3)cc1)O2 \\ \hline
\multirow{3}{*}{Thiothixene} & (canonical) & CN1CCN(CC/C=C2/c3ccccc3Sc3ccc(\textcolor{blue}{S(=O)(=O)N(C)C)cc}32)CC1 \\
                  & rand1 & c1cc2c(cc1)Sc1c(cc(\textcolor{blue}{S(=O)(=O)N(C)C)cc}1)/C2=C\textbackslash\textcolor{red}{CCN1CCN(C)CC}1 \\
                  & rand2 & c1cc2c(cc1)/C(=C/\textcolor{red}{CCN1CCN(C)CC}1)c1cc(\textcolor{blue}{S(=O)(N(C)C)=O)cc}c1S2 \\ \hline
\end{tabular}
\end{table*}

\begin{figure*}[ht]
    \centering
    \includegraphics[width=\linewidth]{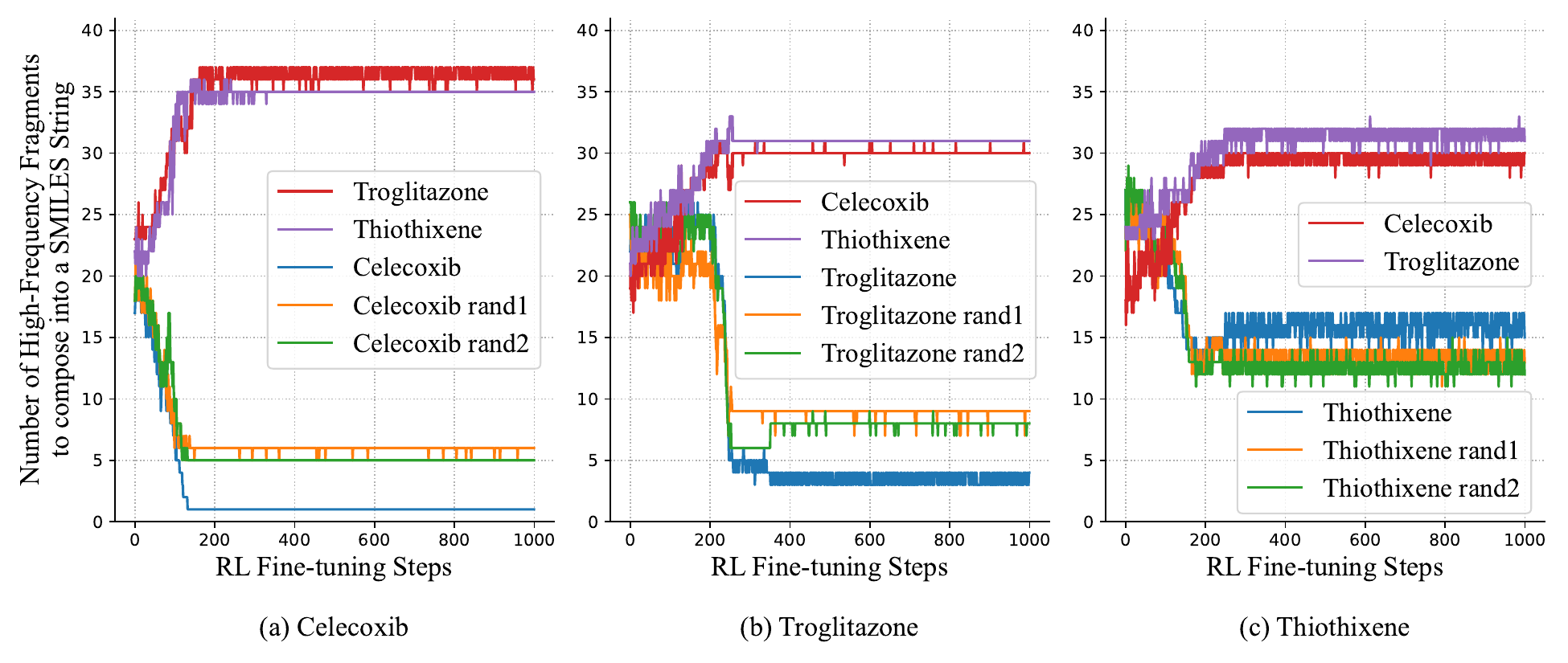}
    \caption{The changing curves of numbers of high-frequency fragments to compose into the given SMILES strings during the RL fine-tuning processes on three drug design tasks.}
    \label{NumFrag}
\end{figure*}

\begin{figure*}[ht]
    \centering
    \includegraphics[width=0.8\linewidth]{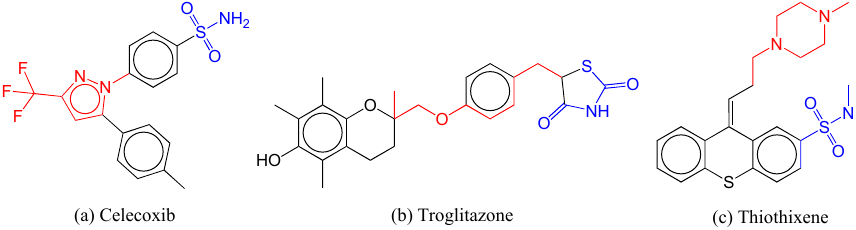}
    \caption{Visualization of substructures in target drug molecules corresponding to high-frequency fragments. The colored substructures correspond to the colored substrings in Table \ref{DrugSmiles}.}
    \label{DrugFrag}
\end{figure*}

\subsection{5.2~~Model's Learning of SMILES Substrings}

According to the experimental results in Section 4.2, RL fine-tuning effectively transfers the language model's chemical knowledge toward the neighborhoods of the drug design objectives. Building on this qualitative conclusion, we aim to further elucidate what kind of knowledge the language model has learned at the SMILES string level that enables it to generate target molecules more effectively. With the tool of the SPE algorithm, we can quantitatively analyze changes in the language model during RL fine-tuning from the perspective of substrings, which aligns with fragment-based strategy in drug design. Our three experimental tasks of drug rediscovery are well-suited for analysis of molecular structure learning, as our learning objectives themselves are structures rather than properties, making it possible to precisely analyze the relationship between the model's knowledge and the structures of the target compounds.

Firstly, we report the quantities of high-frequency substrings extracted by the SPE algorithm from the batch of 256 strings sampled at each RL step during the fine-tuning process of the three tasks; here, the hyper-parameter MF is set to 200. As shown in Figure \ref{highfreq}, the number of high-frequency substrings generated by the pre-trained model (i.e., the agent at the starting point of the fine-tuning process) is approximately between 70 and 80. However, it stabilizes at a higher level with small fluctuations after the fine-tuning converges. This observation is consistent with intuitive cognition: when a language model moves to a small range close to the drug design objective, the chemical knowledge it grasps also focuses on a smaller range. Therefore, the generated sequences should have more identical fragments, i.e., high-frequency substrings.

Next, we select three different SMILES strings for each drug structure, one of which is canonical, while the other two are generated through SMILES randomization, as listed in Table \ref{DrugSmiles}. To demonstrate the quality of task-oriented chemical knowledge learned by the language model, we monitor changes in the efficiency of generating different SMILES representations of the target molecule during the fine-tuning process. This refers to how many high-frequency substrings (as defined above) are needed to construct the target's SMILES string. The smaller this number, the more consistent the chemical spatial structure knowledge mastered by the language model is with the drug design objectives.

Figure \ref{NumFrag} demonstrates the changing curves of numbers of high-frequency fragments to compose into SMILES strings of drug structures during the RL fine-tuning processes. The following conclusions are drawn from Figure \ref{NumFrag}:
\begin{enumerate}
    \item For each task, the number of high-frequency substrings required to constitute different SMILES strings of the target drug molecule converge to a level lower than that of the pre-training model during fine-tuning. This suggests that the language model has acquired knowledge for efficiently generating SMILES strings of the target molecule.
    \item For Celecoxib and Troglitazone, the tasks successfully completed, the numbers of high-frequency substrings needed to constitute the SMILES strings of the target drug molecules are less than that for Thiothixene, a task that does not achieve a full score. Furthermore, fewer high-frequency substrings are required to form canonical SMILES compared to randomized SMILES, which could be related to the pre-training model's better grasp of canonical SMILES.
    \item For molecular structures irrelevant to the task objective (here, we use target molecules of other tasks), the number of high-frequency substrings required to form their SMILES strings converge to a level higher than the pre-training model during fine-tuning. This indicates that the model "forgets" some chemical knowledge unrelated to the task objectives.
\end{enumerate}

\subsection{5.3~~Model's Understanding of Molecular Fragments}

We have pointed out that during the RL fine-tuning process, the language model learns chemical knowledge related to the task objective. This knowledge is reflected in the high-frequency SMILES substrings within the sequences sampled by the model. These substrings both outnumber and are more consistent with the target molecule's SMILES compared to the pre-training model. We aim to further investigate the correspondence between these task-related high-frequency SMILES substrings and the 2D fragments of the target molecules.

In Figure \ref{DrugFrag} and Table \ref{DrugSmiles}, we have color-coded the fragments in the 2D graphs of the target molecules and the corresponding substrings in different SMILES strings. These substrings are all high-frequency ones learned by the GPT agent. From this analysis, we can draw the following conclusions:
\begin{enumerate}
    \item The high-frequency substrings learned by the language model indeed correspond well with the substructures in the 2D molecular graphs, forming wholes in both sequences and the space.
    \item Substrings containing several rare tokens like non-carbon atoms and double bonds are learned by the language model as a whole (even in different sequence forms), and correspond to substructures in the 2D molecular graphs. Examples include \texttt{(S(N)(=O)=O)} in Celecoxib, \texttt{SC(=O)NC3=O} and \texttt{C(=O)NC(=O)S} in Troglitazone, and \texttt{S(=O)(=O)N(C)C)cc} and \texttt{S(=O)(N(C)C)=O)cc} in Thiothixene. These substructures often happen to be key functional groups in drug molecules, playing a decisive role in molecular properties. This indicates that the language model has the potential to learn critical structural features of drug molecules.
    \item In some cases, the 2D molecular fragments do not correspond to contiguous token sequences in SMILES strings. However, this does not prevent the language model from learning corresponding continuous sequences in other SMILES, as exemplified by \texttt{c2cc(C(F)(F)F)nn2-} in Celecoxib (rand1 SMILES) and \texttt{CCN1CCN(C)CC} in Thiothixene (canonical SMILES). This suggests that the language model can overcome the limitations of 1D representations to understand the spatial structure of chemical molecules.
\end{enumerate}

In summary, we believe the language model's understanding of the chemical spatial structure is superior to that at a 1D string level, because it indeed learns structural knowledge related to fragments of target molecules in different tasks, including their key functional groups.

\section{6~~~Conclusion}

In this study, we pre-train and fine-tune a transformer model on SMILES chemical language for drug molecular design. We pay particular attention to the language model's understanding of chemical structures during the fine-tuning process, analyze changes in the quantity and quality of high-frequency SMILES substrings generated by the model, and clarify that the language model indeed understands chemical structures from the correspondence between molecular fragments and these substrings.

Our work lays the foundation for better understanding and designing chemical language models and drug molecular design algorithms. Future research directions include but are not limited to: (1) how the language model understands the relationship between molecular structure and properties; (2) fragment-based language models for drug design; (3) the influence of data on chemical language models.

\bibliography{aaai24}

\end{document}